\documentclass[acmsmall]{acmart}
\AtBeginDocument{%
  }
\usepackage{multirow}
\usepackage{makecell}
\usepackage{booktabs}
\usepackage{array} 
\usepackage{graphicx}
\usepackage{rotating}

\usepackage[caption=false,font=footnotesize,labelfont=rm,textfont=rm]{subfig}
\DeclareSubrefFormat{parens}{#1(#2)}

\usepackage[section]{placeins}
\setcopyright{acmlicensed}
\copyrightyear{2018}
\acmYear{2018}
\acmDOI{XXXXXXX.XXXXXXX}

\acmJournal{JACM}
\acmVolume{37}
\acmNumber{4}
\acmArticle{111}
\acmMonth{8}

\begin{document}

\title{Hierarchical Multimodal LLMs with Semantic Space Alignment
for Enhanced Time Series Classification}

\author{Xiaoyu Tao}
\author{Tingyue Pan}
\author{Mingyue Cheng}
\authornote{Correspondence to: Mingyue Cheng <mycheng@ustc.edu.cn>}
\author{Yucong Luo}
\author{Qi Liu}
\author{Enhong Chen}

\affiliation{
  \institution{State Key Laboratory of Cognitive Intelligence, University of Science and Technology of China}
  \city{Hefei}
  \state{Anhui}
  \country{China}
}

\email{txytiny@mail.ustc.edu.cn}
\email{pty12345@mail.ustc.edu.cn}
\email{mycheng@ustc.edu.cn}
\email{pmri666@mail.ustc.edu.cn}
\email{qiliuql@ustc.edu.cn}
\email{cheneh@ustc.edu.cn}

\renewcommand{\shortauthors}{X. Tao et al.}

\begin{abstract}
  Time series classification plays a fundamental role in a wide range of real-world applications. Recently, large language models (LLMs) have demonstrated strong generalization and reasoning capacities, but directly applying them to time series classification remains non-trivial due to the representation gap between numerical sequences and linguistic semantics. In this paper, we propose HiTime, a hierarchical LLM-based framework for multimodal time series classification that bridges structured temporal representations with semantic reasoning in a generative paradigm. Specifically, we design a hierarchical sequence feature encoding module composed of a data-specific encoder and a task-specific encoder to extract complementary temporal features. To mitigate the embedding gap between time series representations and textual semantics, we further introduce a semantic space alignment module that jointly performs coarse-grained global modeling and fine-grained cross-modal correspondence. Building upon the above representations, we employ a parameter-efficient supervised fine-tuning strategy to activate the generative classification capability of the algined LLMs, thereby transforming conventional discriminative time series classification into a generative task. Extensive experiments on multiple benchmarks demonstrate that the proposed framework consistently outperforms state-of-the-art baselines\footnote{The code is publicly available at https://github.com/Xiaoyu-Tao/HiTime}.
\end{abstract}


\received{20 June 2025}
\received[revised]{20 October 2025}
\received[accepted]{15 November 2025}

\maketitle
\section{Introduction}
Time series classification (TSC) is a fundamental problem in the data mining area and serves as a core technique in a wide range of real-world applications, including healthcare monitoring \citep{zhang2024urban}, industrial diagnosis \citep{zhang2020semi}, human activity recognition \citep{salim2020modelling}, and financial risk assessment \citep{chen2018auto}. Unlike static data, time series are characterized by strong temporal dependencies, non-stationary patterns, and complex dynamic structures, which make accurate classification particularly challenging \citep{fu2016service}. With the rapid growth of sensor systems and multimodal data collection platforms, time series classification is increasingly required to operate under complex, heterogeneous, and semantically rich data environments \citep{liu2019joint, zhang2024towards}. 

Over the past decades, numerous modeling approaches have been developed for time series classification \citep{fu2016service, dempster2021minirocket}. 
Traditional methods include several major categories. 
Distance-based approaches \citep{ding2008querying} rely on elastic similarity measures to compare temporal patterns. 
Shapelet-based approaches \citep{lines2012shapelet} extract discriminative subsequences and use their distances as features for classification.
Dictionary-based approaches \citep{lin2007experiencing} discretize time series into symbolic words and perform classification using frequency-based representations derived from a learned dictionary.
Feature-based approaches \citep{deng2013time} convert time series into feature representations through handcrafted statistical features and classify them with traditional machine learning models. 
Despite their effectiveness, they cannot automatically learn representations from raw time series, a capability that becomes increasingly important as temporal patterns grow more complex. With the advent of deep learning, the focus has shifted to models such as convolutional neural networks (CNNs) \citep{cheng2025convtimenet}, recurrent neural networks (RNNs) \citep{ye2018learning}, graph neural networks (GNNs) \citep{han2021dynamic}, and Transformer-based architectures \citep{zerveas2021transformer}. These models demonstrate superior performance by capturing complex temporal dependencies and patterns inherent in time series, reducing reliance on handcrafted features. 

While deep learning models capture temporal patterns, they lack the generalization and semantic reasoning capability that LLMs provide. Motivated by these strengths, an increasing body of research explores their use in time series analysis \citep{wang2025can,luo2025time}.
These developments can be broadly organized into two main categories. The first category treats LLMs as feature extractors and attaches a linear layer to perform the downstream task. Models such as GPT4TS \citep{zhou2023one} adopts a frozen pre-trained language model as a universal backbone for time series, using it as a feature extractor to achieve strong performance across diverse tasks. 
The second category formulates time series classification as a generative modeling task. Here, temporal sequences are transformed into inputs compatible with LLMs, either through  through discrete tokens \citep{tao2025values} produced by tokenization techniques such as vector quantization (VQ) \citep{van2016wavenet} and k-means \citep{hsu2021hubert} or continuous embeddings \citep{jin2023time} that preserve temporal dynamics. These transformations enable LLMs to model time series within a unified language-based generative framework.

Despite recent advances, existing methods remain suboptimal when directly applying LLMs to time series classification due to the representation gap between numerical sequences and linguistic semantics. Since LLMs are pretrained on large-scale textual corpora \citep{huang2025foundation}, they lack the temporal inductive biases required to capture the structured temporal representations inherent in time series, including periodicity, seasonality, long-term tendencies, and other multi-scale temporal dynamics. Moreover, their limited capability in cross-modal embedding alignment further restricts their ability to associate temporal features with textual semantics \citep{xia2025timeemb}. These limitations collectively impair the model’s capacity to capture the critical structured temporal representations essential for accurate classification, while also preventing LLMs from fully leveraging their semantic reasoning capabilities. Based on the above analysis, there is a clear need for a new approach that can seamlessly integrate temporal representations into the semantic space of LLMs, thereby improving classification performance in diverse practical scenarios.

To address these challenges, we propose \textbf{HiTime}, a hierarchical LLM-based framework for multimodal time series classification, which bridges structured temporal representations and language-based semantic reasoning in a generative paradigm. HiTime starts with a hierarchical sequence feature encoding module composed of a data-specific encoder and a task-specific encoder to extract complementary temporal features. This design allows the framework to preserve general temporal dynamics while remaining adaptable to task-level discriminative patterns. To mitigate the embedding gap between time series representations and textual semantics, we further introduce a semantic space alignment module that jointly performs coarse-grained global modeling and fine-grained cross-modal correspondence. Building upon the above representations, we employ a parameter-efficient supervised fine-tuning strategy to activate the generative classification capability of aligned LLMs, thereby transforming conventional discriminative time series classification into a generative inference process. Through the integration of hierarchical temporal encoding, semantic alignment, and generative instruct fine-tuning, HiTime provides an effective solution for multimodal time series classification, bridging temporal structures with linguistic semantics while leveraging the generalization and semantic reasoning capabilities of LLMs.

Our contribution can be summarized in the following aspects:
\begin{itemize}
    \item We demonstrate that explicitly integrating hierarchical time series feature encoders is critical for enabling effective LLM-based time series classification.
    \item We propose HiTime, a hierarchical LLM-based multimodal framework that integrates data-specific and task-specific temporal embeddings with dual-view semantic alignment.
    \item We conduct comprehensive experiments on public time series datasets to systematically validate the effectiveness of the proposed method.
\end{itemize}

\section{Related Work}
In this section, we first review the comprehensive methods of time series classification and then summarize recent developments that incorporate LLMs into time series analysis.
\subsection{Time Series Classification}
With the growth of sensor-driven applications across healthcare \citep{zheng2024graph}, industrial systems \citep{zheng2014urban}, human activity analysis, and financial risk monitoring \citep{wang2025can}, large volumes of time series are continuously generated. Classifying such data is a fundamental task that attracts extensive attention in the research community \citep{ismail2020inceptiontime}.
Consequently, a variety of modeling approaches emerge, which can be broadly grouped into several major categories \citep{middlehurst2024aeon}. Distance-based methods, such as dynamic time warping (DTW) with k-nearest neighbors \citep{yu2011dynamic}, measure similarity between sequences and handle temporal distortions effectively. Shapelet-based approaches \citep{grabocka2014learning, lines2012shapelet} identify discriminative subsequences that characterize class-specific patterns. Dictionary-based methods, such as symbolic aggregate approximation (SAX) \citep{lin2007experiencing}, transform sequences into symbolic representations for frequency-based analysis. Feature-based techniques, exemplified by time series forest \citep{deng2013time}, extract handcrafted statistical features from predefined intervals.  
Modern deep learning approaches also play a significant role in time series classification. Convolutional neural networks (CNNs) \citep{cheng2025convtimenet} capture hierarchical local temporal patterns through multi-scale convolutional filters. Recurrent neural networks (RNNs) \citep{li2017diffusion, ye2018learning} model sequential dependencies through recurrent connections, while graph neural networks (GNNs) \citep{han2021dynamic, jin2023transferable} incorporate relational structures across multivariate channels. Transformer-based architectures \citep{cheng2023formertime, zerveas2021transformer} leverage self-attention to capture both short- and long-range temporal dependencies. These models automatically learn expressive representations from raw data and reduce reliance on manual feature engineering. However, they typically require substantial labeled data and computational resources, which limits their applicability in low-data or resource-constrained environments.

\subsection{LLM-based Time Series Analysis}
The use of LLMs for time series analysis has attracted increasing research interest \citep{xue2023promptcast, jia2024gpt4mts, liang2024foundation}. Existing approaches can be broadly divided into two major categories.
The first involves integrating LLMs as core components within traditional linear prediction frameworks \citep{liu2024unitime,cao2023tempo,liu2024moirai}. For instance, models like GPT4TS take advantage of the LLM’s sequence modeling strengths to enhance softmax-based classification. While these approaches yield improvements, they do not fully harness the generative potential of LLMs, which could be crucial in  TSC tasks beyond traditional predictive paradigms.
The second approach redefines time series analysis as a generative task within a multimodal understanding framework \citep{pan2024s}. This approach can be further divided into two methodologies. The first adopts discrete token inputs, where techniques are employed to transform time series data into token sequences. These token-based methods leverage natural language processing techniques, allowing LLMs to exploit their generative capabilities fully \citep{ansari2024chronos}. However, despite utilizing LLMs' generative strengths, this approach tends to overlook the temporal dynamics that are intrinsic to time series data.
The second methodology focuses on continuous embeddings, as exemplified by models like Time-LLM \citep{jin2023time}, which re-encode time series data into latent representations to capture temporal dynamics. While this method accounts for the dynamic nature of time series, it does not achieve sufficient semantic alignment between time series and textual features, which limits its effectiveness in multimodal tasks \citep{liu2023unified,chen2024exploring}.
\section{Preliminaries}

To begin with, we present the notations used in this paper, followed by the basic concepts and the formal problem definition. We then provide a brief analysis of the motivation underlying our work.

\subsection{Problem Formulation }
Given a dataset $\mathcal{D} = \{(\mathbf{X}_1, \mathbf{Y}_1), (\mathbf{X}_2, \mathbf{Y}_2), \dots, (\mathbf{X}_N, \mathbf{Y}_N)\}$ of $N$ multivariate time series instances, where each input $\mathbf{X}_i \in \mathbb{R}^{T \times m}$ denotes a time series with $T$ time steps and $m$ observed variables, and $\mathbf{Y}_i$ is the corresponding ground-truth label.
In traditional time series classification, the objective is to learn a function $f_{\text{cls}}: \mathbb{R}^{T \times m} \rightarrow \mathbf{Y}$ that maps each instance $\mathbf{X}_i$ to a discrete class label $\mathbf{Y}_i$.
In contrast, our multimodal classification setting extends the input with additional hybrid information $\mathbf{P}$ and reformulates the task as learning a mapping $f: \{\mathbf{X}, \mathbf{P}\} \rightarrow \mathbf{T}$, where $\mathbf{T}$ is a textual description embedding label-relevant semantics.
Finally, a downstream textual evaluator extracts the predicted class $\hat{\mathbf{Y}}_i$ from $\mathbf{T}_i$, based on criteria such as keyword matching.

\subsection{Motivation Analysis}
LLMs have demonstrated remarkable capabilities in understanding, generating, and reasoning over natural language. Pre-trained on massive text corpora, they possess rich world knowledge and strong generalization abilities, making them powerful tools for a wide range of downstream tasks \cite{zhao2023survey}. Recent efforts have explored extending LLMs to non-text domains, particularly time series analysis and other forms of multimodal learning, motivated by their potential to unify heterogeneous data within a single modeling paradigm. This direction positions LLMs as promising candidates for general-purpose foundation models that integrate natural language information with diverse sensor-driven temporal data \cite{jiang2024empowering}.
However, a key challenge persists: LLMs operate in a discrete semantic space, which is fundamentally misaligned with the continuous and dynamic nature of time series. This embedding gap between time series representations and textual semantics limits their ability to directly interpret temporal patterns or capture structured numerical dynamics, especially in scenarios characterized by multi-source, irregular, and rapidly varying measurements. Bridging this modality mismatch is therefore essential for developing foundation models that can jointly reason over textual knowledge and real-world temporal data, thereby enabling more adaptive, accurate, and context-aware time series classification \cite{jin2023spatio, zhang2022urban, hu2024exploring}.
\section{The Proposed HiTime}
In this section, we present our proposed HiTime framework in detail. We begin with an overview of the overall architecture, followed by the hierarchical feature encoding process. We then describe the semantic space alignment module and the hybrid prompt template. Finally, we outline the generative instruct tuning procedure and the inference strategy.
\begin{figure*}[h]
    \centering
    \includegraphics[width=1\linewidth]{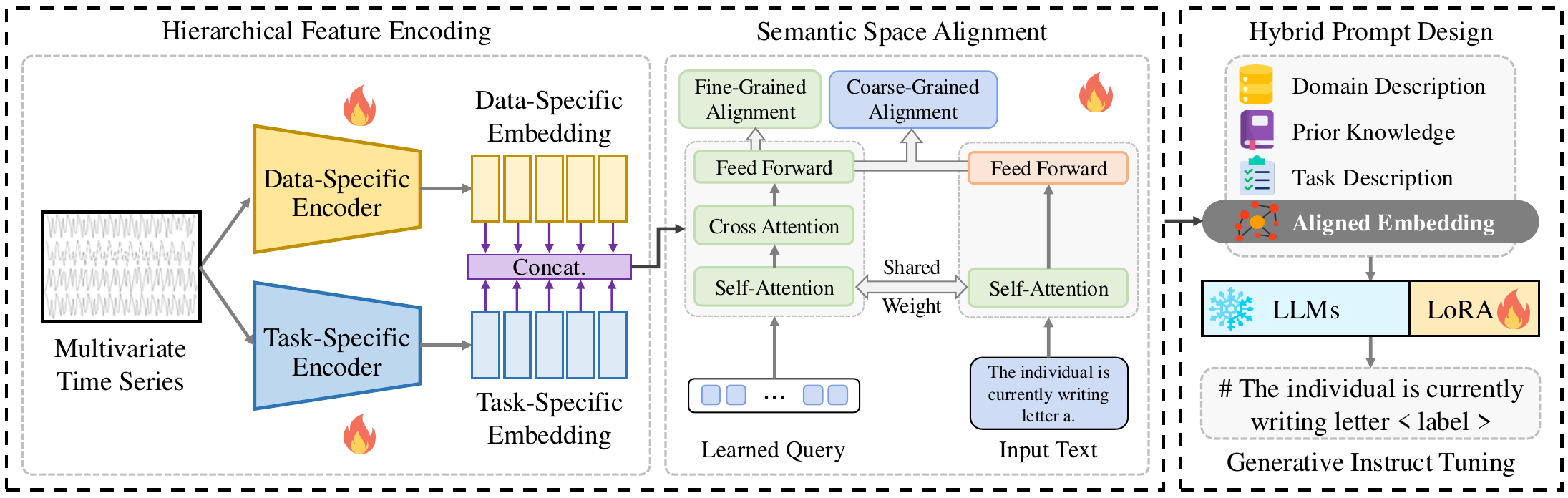}
    \caption{Illustration of the proposed HiTime framework for multimodal time series classification, in which hierarchical feature encoders extract temporal representations, semantic space alignment unifies time–text embeddings, and hybrid prompts enable generative instruct tuning with LLMs. }
    \label{fig:framework}
\end{figure*}
\subsection{Framework Overview}
As shown in Figure \ref{fig:framework}, the overall framework of HiTime consists of three key modules. Given a multivariate time series input, we first process it through hierarchical feature encoding to generate time series embedding that captures both global and local temporal dynamic features. Simultaneously, corresponding textual
information, including class labels, is tokenized to a semantic embedding.  These time series and semantic embeddings are then aligned in a shared latent space, allowing the model to integrate temporal dynamics with textual features. Finally, the aligned embeddings, along with hybrid information, are input into an LLM for generative instruction fine-tuning. This process enhances the model's classification capabilities by leveraging multimodal information and optimizing its understanding of time series data through domain-specific prompts and prior knowledge. In the following sections, we will elaborate on each component in detail.

\subsection{Hierarchical Feature Encoding}
Capturing discriminative dynamic features is essential for time series classification. However, due to their pretraining on large-scale discrete textual data, LLMs inherently lack the inductive bias needed to model temporal variations. To mitigate this limitation, we introduce a hierarchical feature encoding strategy that leverages external lightweight encoders to extract temporal representations from raw time series. Unlike single-task models that often overlook the general properties of temporal data, our approach employs two functionally complementary encoders to jointly model data-specific and task-specific dynamics.

To extract high-quality temporal representations from raw time series, we adopt a hierarchical feature encoding strategy comprising two Transformer-based encoders: a data-specific encoder \citep{cheng2024learning} and a task-specific encoder \citep{cheng2025convtimenet}. The data-specific encoder is trained using a self-supervised learning objective, where 30\% of the input tokens are randomly masked and reconstructed, enabling the model to learn task-relevant dynamics such as short-term fluctuations and long-range dependencies without relying on labels. In contrast, the task-specific encoder is trained in a supervised manner, focusing on learning representations that reflect the intrinsic structure of the data and capture generalizable temporal features. The two types of representations are then concatenated:
\begin{equation}
    \mathbf{E}_c = \operatorname{Concat}(\operatorname{Encoder}_{\text{d}}(\mathbf{X}), \operatorname{Encoder}_{\text{s}}(\mathbf{X})),
    \label{eq:ec}
\end{equation}

where $\mathbf{X}$ is an input instance, $\operatorname{Encoder}_{\text{d}}(\cdot)$ and $\operatorname{Encoder}_{\text{s}}(\cdot)$ are the data-specific and task-specific encoder, $\operatorname{Concat}(\cdot)$ is the concatenation operation, and  $\mathbf{E}_c \in \mathbb{R}^{T \times h}$ denotes the concatenated temporal representation, where $T$ is the sequence length and $h$ is the total feature dimension after concatenating two $h$-dimensional encoder outputs.

By adopting this hierarchical encoding strategy, we convert raw time series into representations that preserve key temporal dynamics as much as possible, complement LLMs’ limitations in temporal modeling, and prepare for subsequent semantic-space alignment.

\subsection{Semantic Space Alignment}
After extracting dynamic temporal representations using hierarchical feature encoders, the next
challenge lies in aligning the representation gap between numerical sequences and linguistic semantics. To this end, we propose a dual-level semantic space alignment module that enables effective matching between the
structured temporal representations and textual semantics. To establish
this alignment, we initialize a set of learnable query tokens $\mathbf{Q} \in \mathbb{R}^{L \times h}$, where $L$ is the number of query vectors and $h$ denotes the hidden feature dimension. These tokens are shared across all instances and act as modality-agnostic anchors for semantic extraction.

The concatenated temporal representation $\mathbf{E}_c \in \mathbb{R}^{T \times h}$, produced by the hierarchical encoder as defined in Eq.~\eqref{eq:ec}, is first attended by the queries via a two-stage interaction:
\begin{align}
    \mathbf{H}_Q &= \mathrm{SelfAttn}(\mathbf{Q}), \\
    \hat{\mathbf{H}}_Q &= \mathrm{CrossAttn}(\mathbf{H}_Q,\ \mathbf{E}_c,\ \mathbf{E}_c),
\end{align}
where $\hat{\mathbf{H}}_Q \in \mathbb{R}^{L \times h}$ is the query output enhanced by temporal context.
Simultaneously, the textual description associated with each instance is embedded as $\mathbf{T} \in \mathbb{R}^{N \times h}$ using a pretrained tokenizer and encoder, and passed through the same self-attention stack:
\begin{equation}
    \mathbf{H}_T = \mathrm{SelfAttn}(\mathbf{T}).
\end{equation}

This shared-weight attention design enforces consistent semantic encoding across modalities, guiding both time series and text representations through comparable transformation patterns and promoting modality-invariant semantics in the latent space. To further reduce the modality gap, we adopt a dual-view optimization strategy with coarse-grained and fine-grained alignment objectives. The coarse-grained objective ensures global semantic consistency, while the fine-grained objective captures local structural correspondence, together producing a more reliably aligned multimodal embedding space. This complementary formulation enables the model to align high-level semantics while remaining sensitive to nuanced variations within temporal patterns.
\paragraph{Coarse-Grained Alignment.}
To align the global semantics between modalities, we formulate a binary classification objective over paired representations. Let \( D = D^+ \cup D^- \) denote the full set of training pairs, where \( D^+ \) is the set of \emph{positive pairs}—time series and text samples from the same class—and \( D^- \) contains \emph{negative pairs}—samples from different classes. Let \( |D| \) denote the total number of all pairs.
We first project the sequence outputs $\hat{\mathbf{H}}_Q$ and $\mathbf{H}_T$ into global embedding vectors $\mathbf{z}_{\mathrm{ts}} \in \mathbb{R}^h$ and $\mathbf{z}_{\mathrm{text}} \in \mathbb{R}^h$ via pooling operations.
The similarity score is computed with a temperature-scaled sigmoid function, providing a smooth probabilistic interpretation of the matching likelihood and stabilizing optimization when embedding magnitudes differ across modalities:

\begin{equation}
S(\mathbf{z}_{\mathrm{ts}}, \mathbf{z}_{\mathrm{text}}) = \sigma\left( \frac{\langle \mathbf{z}_{\mathrm{ts}}, \mathbf{z}_{\mathrm{text}} \rangle}{\tau} \right),
\label{eq:sim_score}
\end{equation}
where \( \langle \cdot, \cdot \rangle \) denotes the inner product, \( \tau \in \mathbb{R}^{+} \) is a learnable temperature scalar, and \( \sigma(\cdot) \) represents the sigmoid activation function applied to normalize the similarity score into the range [0, 1].
The coarse-grained alignment loss $\mathcal{L}_{\text{coarse}}$ is then defined as:

\begin{equation}
\mathcal{L}_{\mathrm{coarse}} = 
- \frac{1}{|D|} \left(
\sum_{(\mathbf{z}_{\mathrm{ts}}, \mathbf{z}_{\mathrm{text}}) \in D^+} \log S(\mathbf{z}_{\mathrm{ts}}, \mathbf{z}_{\mathrm{text}})
+
\sum_{(\mathbf{z}_{\mathrm{ts}}, \mathbf{z}_{\mathrm{text}}) \in D^-} \log (1 - S(\mathbf{z}_{\mathrm{ts}}, \mathbf{z}_{\mathrm{text}}))
\right).
\label{eq:coarse_loss}
\end{equation}

This objective encourages high similarity scores for positive pairs and penalizes negative pairs, thereby guiding the model to learn discriminative global-level alignment between time series and text. This promotes cross-modal representation learning and enhances the model's ability to capture category-level correspondences across modalities.

\paragraph{Fine-Grained Alignment.}
Beyond global semantics, we further incorporate a fine-grained alignment mechanism that explicitly captures detailed correspondences between time series and textual representations. Specifically, we concatenate the temporally-grounded representation $\hat{\mathbf{H}}_Q \in \mathbb{R}^{L \times h}$ with the contextual textual embedding $\mathbf{H}_T \in \mathbb{R}^{T \times h}$ along the sequence dimension. The resulting joint representation is aggregated and subsequently passed through a feed-forward classifier $F_c(\cdot)$ to compute the alignment score, facilitating precise semantic interaction across modalities:

\begin{equation}
\hat{y} = F_c\left( \text{Concat}\left( \hat{\mathbf{H}}_Q, \mathbf{H}_T \right) \right),
\end{equation}
where $\hat{\mathbf{H}}_Q$ and $\mathbf{H}_T$ denote the global representations of the time series and text, respectively, and $F_c(\cdot)$ is a feed-forward classifier that outputs an alignment score $\hat{y} \in [0,1]$.
To train the alignment module, we adopt a binary cross-entropy loss defined over a set of instance pairs $\mathcal{D}$, where each pair $(x, t)$ consists of a time series instance and its corresponding textual description. This fine-grained alignment loss $\mathcal{L}_{\text{fine}}$ guides the model to assign high scores to matched pairs:
\begin{equation}
\mathcal{L}_{\text{fine}} = - \frac{1}{|\mathcal{D}|} \sum_{(x, t) \in \mathcal{D}} \left( y \log \hat{y} + (1 - y) \log (1 - \hat{y}) \right),
\end{equation}
where $y \in \{0,1\}$ denotes the ground-truth alignment label indicating whether the pair $(x, t)$ is positively or negatively matched, and $\hat{y}$ is the predicted alignment probability obtained from the classifier $F_c(\cdot)$ applied to the concatenated joint representation.

\paragraph{Training Objective.}
The overall alignment objective  $\mathcal{L}_{\text{align}}$ combines the two alignment views:
\begin{equation}
    \mathcal{L}_{\text{align}} = \alpha \cdot \mathcal{L}_{\text{coarse}} + \beta \cdot \mathcal{L}_{\text{fine}},
\end{equation}
where $\alpha, \beta \in \mathbb{R}^{+}$ are weighting coefficients that control the trade-off between the coarse-grained and fine-grained alignment losses. In our implementation, we set both coefficients to the same value, thereby assigning equal importance to both levels of alignment during training. This balanced weighting strategy enables the model to jointly capture high-level semantic consistency and fine-grained token correspondences across modalities. After the alignment process, the representations from both modalities are projected into a shared latent space, yielding enriched and semantically aligned embeddings for downstream tasks.
Our training paradigm is inspired by the overall optimization strategy in BLIP-2~\citep{li2023blip}, which provides an effective foundation for multimodal alignment. Rather than directly adopting the original framework, we adapt its dual-level semantic alignment module to the characteristics of time series classification, where dynamic patterns and temporal dependencies are central. Due to space constraints, we omit a detailed review of BLIP-2 and refer readers to the original literature for further details.

\begin{figure*}[ht]
    \centering
    \includegraphics[width=0.85\linewidth]{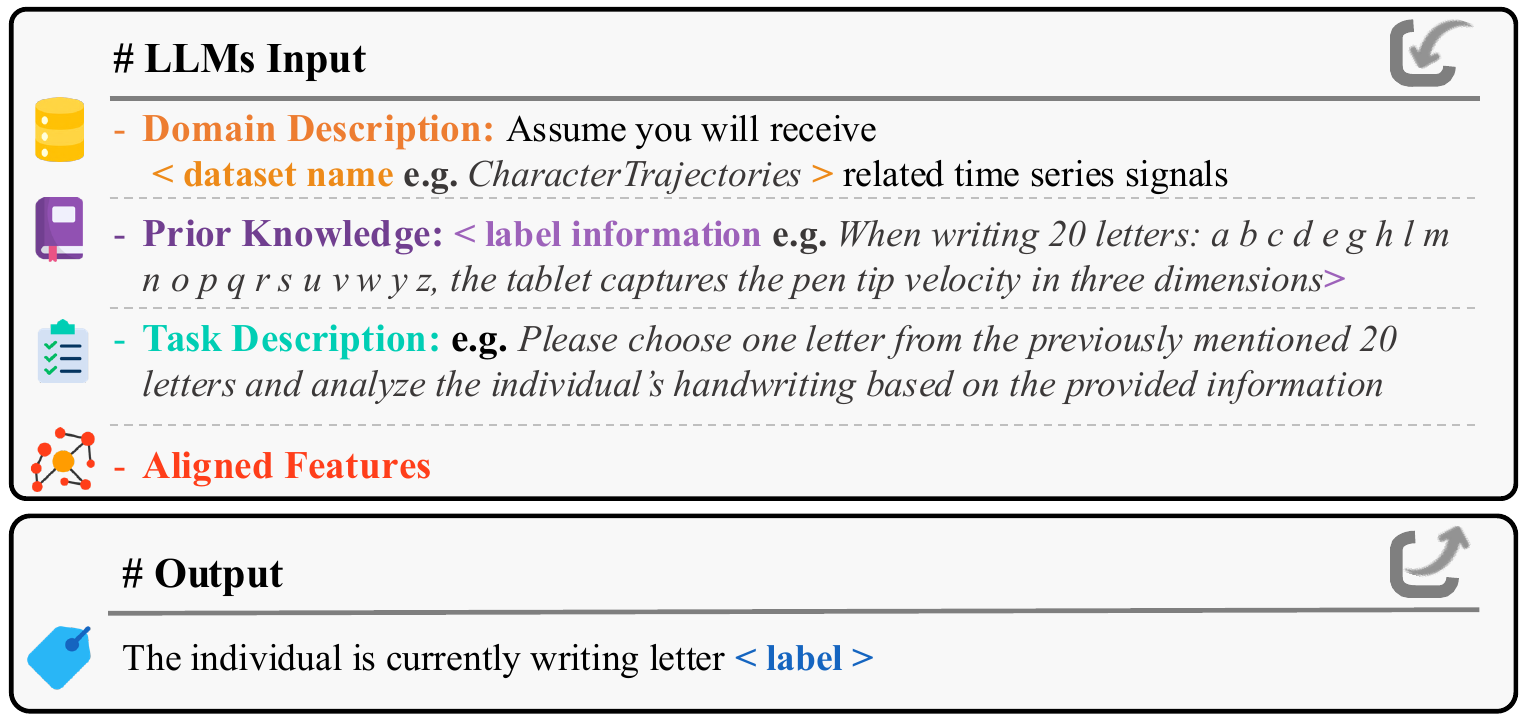}
    \caption{Illustration of the designed hybrid prompt template in HiTime.}
    \label{fig:prompt_tem}
\end{figure*}
\subsection{Hybrid Prompt Template}
Task instructions can effectively guide LLMs to switch to the corresponding tasks to perform the required operations \citep{liu2024visual}. Therefore, the design of prompts plays a vital role in language generation tasks. Well-structured and high-quality prompts not only help the model better comprehend task requirements but also provide essential context and direction, ensuring that the generated output aligns with expectations.  For each domain, we employ a fixed and hybrid prompt to generate target labels, as illustrated in Figure \ref{fig:prompt_tem}. 
This prompt consists of four key components: domain description, prior knowledge, task description, and aligned embedding. These components form a unified yet adaptive structure that allows the prompt to generalize across heterogeneous domains. The domain description provides essential background information about the input time series, helping the model recognize unique characteristics and data distributions specific to each dataset. Prior knowledge introduces domain-relevant expertise, enabling the model to perform accurate reasoning for complex forecasting and classification tasks. The task description serves as a detailed guideline, ensuring that the LLM processes time series data in strict accordance with task requirements. The aligned embedding integrates time series and textual representations within a shared semantic space. During training, only the textual content within these components is updated to reflect the characteristics of each dataset, while the overall prompt structure and alignment mechanism remain consistent. This design allows the model to preserve a unified reasoning framework while flexibly adapting to domain-specific contexts.

\subsection{ Generative Instruct Tuning and Inference}
Our training process and generative classification procedure will be analyzed in detail here. 
The focus of training our framework lies in these modules: the hierarchical feature encoder, the semantic space alignment module, and the LoRA \citep{hu2021lora} module. The objective of fine-tuning the hierarchical feature encoders is to enhance the extraction of high-quality dynamic features from time series data. The primary goal of fine-tuning the semantic space alignment module is to foster interaction and mutual understanding between time series and text features, effectively bringing data from different modalities into the same semantic space. 
The fine-tuning of the LoRA module focuses on utilizing time series features for classification tasks. As a parameter-efficient fine-tuning method, LoRA updates only a small subset of weights, offering a computationally efficient alternative to expensive full-parameter tuning while keeping the LLM centered on generative modeling. During this stage, next-token prediction loss guides the model to generate accurate classification labels based on time series features. This enables LoRA to adjust targeted parameters so that the model can effectively produce classification results through text generation.
In the inference stage, we adopt a generative classification approach, where the model generates textual outputs from which the final label is derived. The key distinction enabling this paradigm lies in the qualitative leap from traditional language models (LMs) to modern LLMs. While conventional LMs typically support shallow text processing and simple label prediction, LLMs—with substantially larger parameter capacity—possess stronger semantic reasoning abilities and a richer internal knowledge base. This allows them to interpret prompts more deeply, integrate contextual cues from both textual instructions and time-series tokens, and consequently generate more accurate and semantically coherent outputs.
After generation, we apply keyword-based extraction techniques to ground the generated text into discrete labels. This strategy leverages the advanced generative and semantic reasoning capabilities of LLMs while maintaining a lightweight and reliable post-processing mechanism. The ability of LLMs to produce detailed, semantically enriched text marks a fundamental improvement over traditional LMs, enabling more precise and context-aware classification. This qualitative enhancement in model capability is central to the effectiveness and adaptability of our generative classification framework.

\section{Experiments}
In this section, we first present the experimental settings used to evaluate HiTime. We then analyze its classification performance, followed by ablation studies that examine the contributions of key modules. Next, we provide an in-depth analysis to further investigate the behavior of HiTime. Finally, we illustrate representative examples through case studies.

\begin{table}[h]
  \centering
  \caption{Statistics of the multivariate time series datasets used in our experiments.}
    \begin{tabular}{cccccc}
    \toprule
    Datasets & \#Train & \#Test & \#Channels & \#Sequence Length & \#Classes \\
    \midrule
    CT    & 1,422  & 1,436  & 3     & 182   & 20 \\
    EP    & 137   & 138   & 3     & 206   & 4 \\
    FD    & 5,890  & 3,524  & 144   & 62    & 2 \\
    HB    & 204   & 205   & 61    & 405   & 2 \\
    NAT & 180   & 108   & 24    & 51    & 6 \\
    PD    & 7,494  & 3,496  & 2   & 8   & 10 \\
    PEM & 267   & 173   & 963   & 144   & 7 \\
    SAD   & 6,599  & 2,199  & 13    & 93    & 10 \\
    SRS1  & 268   & 293   & 6     & 896   & 2 \\
    SRS2  & 200   & 180   & 7     & 1,152  & 2 \\
    \bottomrule
    \end{tabular}%
  \label{tab:clasification_dataset}%
\end{table}%

\subsection{Experimental Settings}
\subsubsection{Datasets.}

We conduct experiments on ten publicly available datasets from the UEA multivariate time series classification archive \citep{bagnall2018uea}, a widely adopted benchmark suite for evaluating multivariate sequence learning models. These datasets span diverse domains, including motion capture, brain–computer interfaces, handwriting recognition, and traffic monitoring. Table \ref{tab:clasification_dataset} summarizes their key properties, including sequence length, number of variables, and number of categories.
CharacterTrajectories (CT): Pen-tip trajectories that simulate handwriting behaviors in interactive or educational applications.
Epilepsy (EP): EEG signals recorded from wearable devices, commonly used in clinical monitoring and neurological studies.
FaceDetection (FD): Facial motion features extracted from videos, relevant to perception and interaction systems.
Heartwriting (HB): Pen-drawn digit traces captured by smart pens, supporting studies on fine-grained motion patterns.
NATOPS (NAT): Hand gesture data collected from motion sensors, applicable to human–machine interaction and gesture-control tasks.
PenDigits (PD): Stylus-based digit-writing sequences used for handwriting recognition and digital input modeling.
PEMS-SF (PEM): Traffic sensor readings from California freeways, widely used in mobility and transportation prediction research.
SpokenArabicDigits (SAD): Spoken digit sequences supporting speech-based classification and recognition applications.
SelfRegulationSCP1/2 (SRS1, SRS2): Electroencephalography recordings collected during cognitive tasks, commonly used to analyze brain dynamics and adaptive interface behaviors.

\subsubsection{Baselines.}
To evaluate the effectiveness of our proposed HiTime framework on time series classification tasks, we compare it against a broad set of state-of-the-art baselines from diverse methodological categories, including distance-based models, convolutional and transformer architectures, as well as LLM-based approaches. These models are widely applied to various real-world time series tasks, many of which originate from urban scenarios, such as traffic flow analysis, energy consumption monitoring, and human activity recognition in smart environments.
NN-DTW \citep{ding2008querying} uses dynamic time warping together with nearest-neighbor voting to classify sequences based on temporal similarity, particularly effective when variable pacing and alignment are important. Shapelet Transformer (ST) \citep{lines2012shapelet} identifies discriminative sub-sequences (shapelets) and uses them as features, providing interpretable representations for time series classification. MCDCNN \citep{zheng2014time} and MCNN \citep{cui2016multi} employ convolutional layers to capture local and multi-scale temporal patterns, enabling hierarchical feature extraction from multivariate sequences. MiniRocket \citep{dempster2021minirocket} offers an efficient feature-extraction mechanism based on random convolutional kernels, achieving strong performance with significantly reduced computational cost. TST \citep{zerveas2021transformer} and Crossformer \citep{zhang2023crossformer} extend Transformer architectures to time series by capturing global dependencies and modeling interactions across variables. Reformer \citep{kitaev2020reformer} introduces memory-efficient attention mechanisms, enabling long-sequence modeling with reduced computational overhead. TimesNet \citep{wu2022timesnet} leverages spectral decomposition and 2D convolutions to extract periodic and structural information from time series. DLinear \citep{zeng2023transformers} employs simple seasonal–trend decomposition with linear modeling and achieves competitive performance across diverse benchmarks.
We also include LLM-based baselines designed for time series. GPT4TS \citep{zhou2023one} reformats time series for GPT-2-based sequence modeling, enabling a frozen transformer to operate directly on temporal data. CrossTimeNet \citep{cheng2024learning} learns transferable representations through cross-domain self-supervised pre-training with VQ-based time series tokenization. Time-LLM \citep{jin2023time} reprograms LLMs using natural-language prototypes, enabling strong few-shot and zero-shot performance. Time-FFM \citep{liu2024time} incorporates federated training and dynamic prompt adaptation for personalized modeling across distributed data sources.

\begin{figure}[t]
    \centering
    \includegraphics[width=0.75\linewidth]{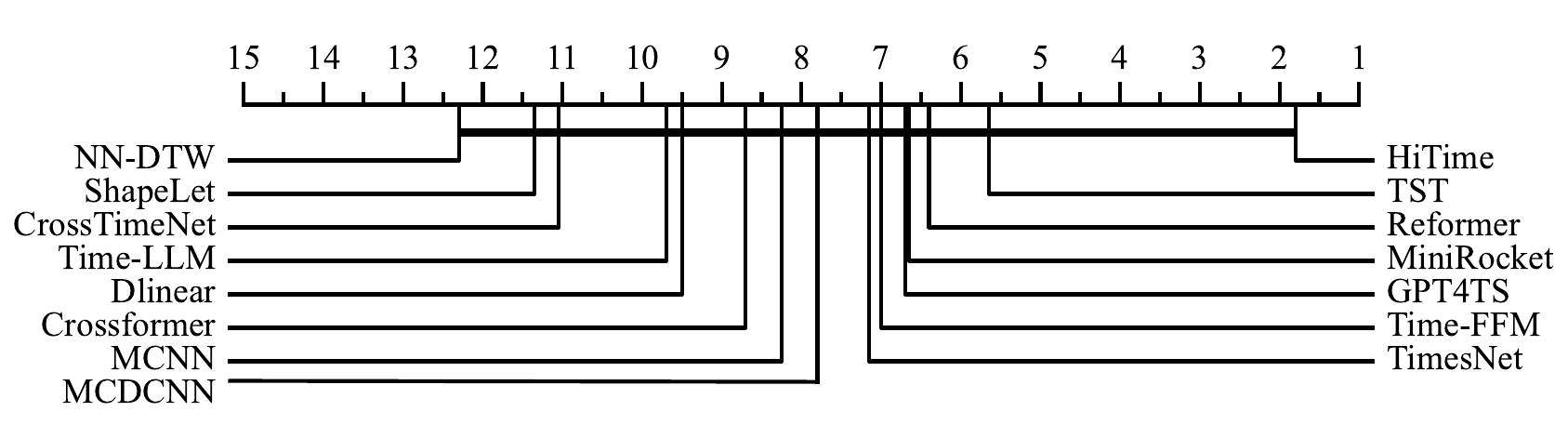}
    \caption{Critical difference diagram over the mean ranks of HiTime, baseline methods.}
    \label{fig:cd-diagram}
\end{figure}

\subsubsection{Implementation Details}

This experiment consists of three steps, each training a different module. Step 1 mainly trains the time series feature extraction encoder. We utilize CrossTimeNet\cite{cheng2024learning} as our data-specific encoder and ConvTimeNet \cite{cheng2025convtimenet} as our task-specific encoder. To achieve data balance, the data is divided according to the size of the dataset. Specifically, the dataset is divided into two categories: small-scale (EP, HB, NAT, PEM, SRS1, SRS2) and large-scale (CT, FD, PD, SAD) for general encoder training. The data-specific encoder is configured with a hidden dimension of 1024, a learning rate of 0.0001, and a dropout rate of 0.1. For the task-specific encoder, ConvTimeNet, the deep-wise convolution kernel sizes are chosen from \{[7,7,13,13,19,19], [19,19,29,29,37,37], [37,37,43,43,53,53]\},  and the Patch Size is evaluated over \{8, 16, 32\}. In step 2, the alignment module is pre-trained with a learning rate of 0.001 and a dropout rate of 0.1 to ensure effective pre-training of the alignment module. In step 3, supervised fine-tuning (SFT) is performed on LLMs using LoRA technology. The rank of LoRA is chosen from \{32, 16, 8\}, and the LoRA alpha is evaluated over \{16, 8\}, the warmup step is selected from \{1,000, 5,000\}, the dropout rate is chosen from \{0.05, 0.1\}, the initial learning rate is set to 0.0001, and the warmup learning rate is set to 1e-5. The optimizer is AdamW, and different schedulers are used in different stages: the LambdaLR scheduler is used in step 1, and the CosineLR scheduler is used in step 2 and step 3 to ensure effective training and fine-tuning. To be fair, we replace the backbone of Time-LLM and Time-FFM with the currently more advanced public model Llama3.1-8B. For all experiments, we use accuracy and F1 score as the primary evaluation metrics for classification performance, with bold indicating the best results and underline marking the second-best results.

\begin{table*}[htbp]
  \centering
  \caption{Main results for classification evaluated by accuracy. }
  \resizebox{\textwidth}{!}{
    \begin{tabular}{c|ccccccccccc}
    \toprule
    Models & CT    & EP    & FD    & HB    & NAT & PD    & PEM & SAD   & SRS1  & SRS2  & Average \\
    \midrule
    ST & 0.9722  & 0.8019  & 0.5677  & 0.7398  & 0.8296  & 0.9809  & 0.6069  & 0.9847  & 0.8020  & 0.5648  & 0.7851  \\
    NN-DTW & 0.9903  & 0.8913  & 0.5329  & 0.6732  & 0.8667  & 0.9774  & 0.6936  & 0.9595  & 0.7850  & 0.4833  & 0.7853  \\
    MCDCNN & 0.9858  & 0.9638  & 0.5969  & 0.7561  & 0.9074  & 0.9805  & 0.7803  & 0.9815  & 0.9010  & 0.5278  & 0.8381  \\
    MCNN  & 0.9861  & 0.9614  & 0.6613  & 0.7220  & 0.8667  & 0.9661  & 0.7631  & 0.9888  & 0.9113  & 0.5352  & 0.8362  \\
    ConvTimeNet  & \underline{0.9937} & \textbf{0.9855}  & 0.6804  & 0.7301  & \underline{0.9667}  & 0.9814  & 0.8245  & \underline{0.9916}  & 0.8907  & 0.5556  & \underline{0.8600}  \\
    MiniRocket & 0.9817  & \underline{0.9819}  & 0.6317  & \underline{0.7598}  & 0.8759  & \underline{0.9847}  & 0.7206  & 0.9891  & 0.8458  & 0.5648  & 0.8336  \\
    TimesNet & 0.9761  & 0.8913  & 0.6902  & 0.7285  & 0.8907  & 0.9809  & 0.8537  & 0.9898  & 0.9192  & 0.5185  & 0.8439  \\
    TST   & 0.9833  & 0.8889  & 0.6897  & 0.7463  & 0.9463  & 0.9820  & 0.8420  & 0.9862  & 0.9124  & 0.5704  & 0.8548  \\
    Crossformer & 0.9759  & 0.8077  & 0.6927  & 0.7236  & 0.9352  & 0.9796  & 0.8092  & 0.9785  & 0.9022  & 0.5352  & 0.8340  \\
    Reformer & 0.9821  & 0.8527  & \textbf{0.6936 } & 0.7220  & 0.9241  & \textbf{0.9859 } & 0.8314  & 0.9861  & \underline{0.9206}  & 0.5537  & 0.8452  \\
    DLinear & 0.9759  & 0.6087  & \underline{0.6934}  & 0.7398  & 0.9278  & 0.8685  & 0.8651  & 0.9707  & 0.8805  & 0.5111  & 0.8042  \\
    GPT4TS & 0.9844  & 0.7440  & 0.6916  & 0.7220  & 0.8926  & 0.9827  & \underline{0.8726}  & 0.9877  & 0.9101  & 0.5685  & 0.8356  \\
    CrossTimeNet & 0.9247  & 0.9130  & 0.5808  & 0.7463  & 0.7666  & 0.9645  & 0.6820  & 0.9690  & 0.8328  & 0.5709  & 0.7951  \\
    Time-LLM & 0.9865  & 0.9330  & 0.6003  & 0.7220  & 0.8056  & 0.9783  & 0.7271  & 0.9811  & 0.8123  & 0.5667  & 0.8015
 \\
    Time-FFM & 0.9923  & 0.9638  & 0.5900  & 0.7220  & \underline{0.9667}  & 0.9811  & 0.7052  & 0.9891  & 0.8191  & \underline{0.5722}  & 0.8145
 \\
 \midrule
    HiTime  & \textbf{0.9944 } & \textbf{0.9855 } & 0.6771  & \textbf{0.7610 } & \textbf{0.9833 } & 0.9824  & \textbf{0.8844 } & \textbf{0.9932 } & \textbf{0.9249 } & \textbf{0.5778 } & \textbf{0.8764 } \\
    \bottomrule
    \end{tabular}%
     }
  \label{tab:main_result}%
\end{table*}%
\begin{table*}[htbp]
  \centering
  \caption{Main results for classification evaluated by F1 score. }
  \resizebox{\textwidth}{!}{%
    \begin{tabular}{c|ccccccccccc}
    \toprule
    Models & CT    & EP    & FD    & HB    & NAT & PD    & PEM & SAD   & SRS1  & SRS2  & Average \\
    \midrule
    Shapelet & 0.9715  & 0.7869  & 0.5613  & 0.5557  & 0.8289  & 0.9810  & 0.5975  & 0.9847  & 0.7995  & 0.5214  & 0.7588  \\
    NN-DTW & 0.9894  & 0.8890  & 0.5329  & 0.5711  & 0.8684  & 0.9775  & 0.6842  & 0.9597  & 0.7800  & 0.4787  & 0.7731  \\
    MCDCNN & 0.9853  & 0.9637  & 0.5411  & 0.5986  & 0.9080  & 0.9805  & 0.7716  & 0.9815  & 0.9007  & 0.5053  & 0.8136  \\
    MCNN  & 0.9852  & 0.9592  & 0.6602  & 0.4193  & 0.8662  & 0.9664  & 0.7401  & 0.9888  & 0.9111  & 0.5130  & 0.8010  \\
    MiniRocket & 0.9808  & \underline{0.9812 } & 0.6308  & \textbf{0.6890 } & 0.8758  & \underline{0.9846 } & 0.7082  & 0.9891  & 0.9203  & 0.5490  & 0.8309  \\
    TimesNet & 0.9742  & 0.8809  & 0.6900  & 0.4513  & 0.8768  & 0.9816  & 0.8491  & \underline{0.9898 } & 0.9192  & 0.4351  & 0.8048  \\
    TST   & 0.9822  & 0.8808  & 0.6895  & 0.6083  & 0.9462  & 0.9821  & 0.8359  & 0.9862  & 0.9123  & 0.5593  & \underline{0.8383}  \\
    Crossformer & 0.9743  & 0.7910  & 0.6923  & 0.4306  & 0.9367  & 0.9797  & 0.8023  & 0.9785  & 0.9021  & 0.4562  & 0.7944  \\
    Reformer & 0.9813  & 0.8412  & \underline{0.6932 } & 0.4193  & 0.9261  & \textbf{0.9859} & 0.8215  & 0.9861  & \underline{0.9206 } & 0.5378  & 0.8113  \\
    DLinear & 0.9744  & 0.5596  & \textbf{0.6933 } & 0.5558  & 0.9282  & 0.8677  & 0.8616  & 0.9708  & 0.8805  & 0.4192  & 0.7711  \\
    GPT4TS & 0.9843  & 0.7331  & 0.6903  & 0.4193  & 0.8870  & 0.9828  & \underline{0.8709 } & 0.9877  & 0.9099  & 0.5474  & 0.8013  \\
    CrossTimeNet & 0.9131  & 0.9047  & 0.5716  & 0.6091  & 0.7578  & 0.9613  & 0.6791  & 0.9703  & 0.8297  & 0.5677  & 0.7764  \\
    Time-LLM & 0.9861  & 0.9216  & 0.5740  & 0.4193  & 0.8079  & 0.9714  & 0.7019   & 0.9824  & 0.8108  & 0.5662  & 0.7822  \\
    Time-FFM & \underline{0.9915}  & 0.9637  & 0.5899  & 0.4193  & \underline{0.9667}  & 0.9812  & 0.6990  & 0.9891  & 0.8191  & \underline{0.5700}  & 0.7989  \\
    \midrule
    HiTime & \textbf{0.9941 } & \textbf{0.9850 } & 0.6768  & \underline{0.6231 } & \textbf{0.9833 } & 0.9830  & \textbf{0.8811 } & \textbf{0.9932 } & \textbf{0.9249 } & \textbf{0.5744 } & \textbf{0.8619 } \\
    \bottomrule
    \end{tabular}%
    }
  \label{tab:main_f1}%
\end{table*}%
\subsection{Classification Performance Analysis}

Tables \ref{tab:main_result} and \ref{tab:main_f1} provide a comprehensive comparison of classification results across all methods in terms of accuracy and F1 score, whereas Figure \ref{fig:cd-diagram} illustrates the corresponding critical difference diagram following the statistical procedure in \citep{demvsar2006statistical}. Compared to other baseline models, HiTime consistently delivers either the best or near-optimal classification performance, with its average performance standing out prominently. This provides evidence of its effectiveness in enhancing the TSC task. Notably, no single model can achieve optimal performance across all scenarios \citep{ruiz2021great,bagnall2017great}, making HiTime’s ability to maintain an overall leading performance a particularly strong result. 

From the experimental results in Table \ref{tab:main_result}, several key observations can be made. First, HiTime demonstrates the highest performance among all LLM- and LM-based methods. This superior performance can be attributed to its hierarchical feature encoding strategy, which bridges structured temporal representations with the semantic reasoning capabilities of LLMs. This effect is especially beneficial for datasets like NAT, where capturing both semantic and temporal dependencies is crucial. Second, HiTime consistently outperforms Transformer-based methods across most datasets, notably excelling on the PEM dataset. This advantage can likely be ascribed to HiTime’s ability to leverage LLMs' pre-trained knowledge, enhancing its reasoning capability for time series data. By integrating global contextual understanding, HiTime achieves more effective feature learning than traditional Transformer architectures. Based on the results presented in Table \ref{tab:main_f1}, the proposed HiTime model clearly establishes state-of-the-art performance by achieving the highest average F1 score, demonstrating a superior comprehensive effectiveness that surpasses all other baselines and remains stable across diverse scenarios, further highlighting its adaptability in practical applications. 

However, on the PD and FD datasets, HiTime underperforms compared to Reformer.
The performance discrepancy may stem from differences in data complexity
and feature regularity. PD consists of highly structured, low-variance trajectories where convolutional
baselines better exploit repetitive local patterns, while FD contains noisy, high-dimensional motion
features that challenge temporal alignment. From a model perspective, HiTime prioritizes semantic
reasoning and long-range contextual integration, which benefits complex multimodal tasks but may
underutilize fine-grained cues in simpler or noisier datasets. These results highlight the inherent trade-off
between global semantic modeling and local pattern sensitivity, and we acknowledge that no single model can handle all types of time-series tasks.

\begin{table*}[htbp]
  \centering
  
  \caption{ Performance comparison of different backbone models.}
  \resizebox{\textwidth}{!}{
    \begin{tabular}{c|ccccccccccc}
    \toprule
    Backbones & CT    & EP    & FD    & HB    & NAT   & PD    & PEM   & SAD   & SRS1  & SRS2  & \multicolumn{1}{c}{Average} \\
    \midrule
    GPT-2  & \underline{0.9903}  & \textbf{0.9928 } & 0.6568  & \underline{0.7220}  & 0.9667  & 0.9760  & 0.8439  & \textbf{0.9977 } & 0.8976  & 0.5000  & \underline{0.8544}  \\
    Llama3.2-1B & 0.9882 & 0.9710 & 0.6669 & 0.6951 & 0.9644 & \underline{0.9817} & 0.8550 & 0.9899 & \underline{0.9042} & \underline{0.5225} & 0.8539\\
    Llama3.2-3B  & 0.9887 & 0.9710 & \underline{0.6688} & 0.7127 & \underline{0.9733} & 0.9748 & \underline{0.8666} & 0.9864 & 0.8901 & 0.5114 & \underline{0.8544}\\
    Llama3.1-8B & \textbf{0.9944 } & \underline{0.9855}  & \textbf{0.6771 } & \textbf{0.7610 } & \textbf{0.9833 } & \textbf{0.9828 } & \textbf{0.8844 } & \underline{0.9932}  & \textbf{0.9249 } & \textbf{0.5778 } & \multicolumn{1}{c}{\textbf{0.8764 }} \\
    \bottomrule
    \end{tabular}%
    }
  \label{tab:model_comparison}%
\end{table*}%
\subsection{Ablation Studies of Key Modules}
\subsubsection{Performance of Backbone Models.} To evaluate the influence of different backbone architectures, we compare HiTime constructed on four models of varying parameter scales across ten datasets. As shown in Table~\ref{tab:model_comparison}, the Llama3.1-8B backbone achieves the highest overall performance, with an average score of 0.8764, indicating that increased model capacity generally enhances representational strength and semantic–temporal alignment. A closer inspection of dataset-level results further reveals that larger backbones yield substantial improvements on tasks requiring richer temporal–semantic reasoning, such as CT, HB, NAT, and PEM. GPT-2, although competitive on EP and SAD, exhibits clear limitations on more complex datasets like SRS1 and SRS2, reflecting its reduced ability to capture long-range temporal dependencies. In contrast, Llama3.1-8B attains the best performance on nine out of ten datasets, demonstrating strong stability across heterogeneous time series. These findings suggest that larger backbones better integrate hierarchical temporal features with textual semantics, reduce sensitivity to dataset-specific patterns, and enable HiTime to maintain stable performance across tasks. Overall, the results indicate that backbone scalability is central to HiTime’s classification effectiveness.

\begin{table}[h]
  \centering
  \caption{Experimental results of hierarchical feature encoders. “w/o encoder” indicates a model variant in which feature encoding modules are entirely removed.}
  \resizebox{0.85\textwidth}{!}{
    \begin{tabular}{ccccc}
    \toprule
    Datasets & Ours & Task-specific Encoder & Data-specific Encoder & w/o Encoder \\
    \midrule
    CT    & \textbf{0.9944 } & \underline{0.9875}  & 0.9802 & 0.8496  \\
    EP    & \textbf{0.9855 } & \underline{0.9838}  & 0.9203 & 0.9130 \\
    FD    & \textbf{0.6771 } & 0.6643  & \underline{0.6683} & 0.5000 \\
    HB    & \textbf{0.7610 } & 0.7195  & \underline{0.7317} & 0.6951 \\
    NAT & \textbf{0.9833 } & 0.9222  & 0.9222 & \underline{0.9370} \\
    PD    & \textbf{0.9828 } & 0.9817  & \underline{0.9820} & 0.7098 \\
    PEM & \textbf{0.8844 } & \underline{0.8497}  & \underline{0.8497} &  0.7866\\
    SAD   & \textbf{0.9932 } & \textbf{0.9932 } & \underline{0.9914} &  0.9150 \\
    SRS1  & \textbf{0.9249 } & \underline{0.8601}  & 0.8362 & 0.8532 \\
    SRS2  & \textbf{0.5778 } & \underline{0.5467}  & 0.5111 & 0.4944 \\
    \midrule
    Average & \textbf{0.8764}  & \underline{0.8509}  & 0.8393 & 0.7654 \\
    \bottomrule
    \end{tabular}%
    }
  \label{tab:encoding}%
\end{table}%
\subsubsection{Impact of Hierarchical Feature Encoder.}

To explore the effectiveness of the proposed hierarchical feature encoder, four temporal encoding variants are designed and tested on ten datasets. As shown in Table \ref{tab:encoding}, the hierarchical feature encoder consistently achieves superior performance across most datasets. This result demonstrates that hierarchical encoding effectively models time series data by integrating task-specific features with intrinsic data characteristics, thereby further enhancing overall classification accuracy in practical settings.
In contrast, the data-specific encoder underperforms, likely due to its limited adaptability and insufficient feature extraction accuracy, making it less effective at capturing key patterns in complex and dynamic time series. Additionally, the model without feature encoding exhibits the weakest overall performance, highlighting the critical role of structured temporal representations in improving classification accuracy.
Notably, on the NAT dataset, both data-specific and task-specific encodings fail to achieve optimal results individually. However, the hierarchical feature encoder, which integrates the strengths of both, significantly surpasses the model without feature encoding. This finding underscores the importance of simultaneously modeling task-related features and intrinsic data characteristics for reliable time series classification performance across datasets.

\begin{table}[ht]
  \centering
  \caption{Experimental results of different alignment strategies.}
  \resizebox{1\textwidth}{!}{
    \begin{tabular}{ccccc}
    \toprule
    Datasets & Coarse-Grained and Fine-Grained Alignment & Fine-Grained Alignment   & Coarse-Grained Alignment   & Random Alignment\\
    \midrule
    CT    & \textbf{0.9944 } & \underline{0.9916}  & 0.9798  & \underline{0.9916}  \\
    EP    & \textbf{0.9855 } & \underline{0.9673}  & \textbf{0.9855 } & 0.9493  \\
    FD    & \textbf{0.6771 } & 0.6555  & \underline{0.6720}  & 0.6694  \\
    HB    & \textbf{0.7610 } & \underline{0.7366}  & 0.7317  & 0.7024  \\
    NAT & \textbf{0.9833 } & \underline{0.9778}  & 0.9467  & 0.9389  \\
    PD    & \underline{0.9828}  & 0.9808  & \textbf{0.9848 } & 0.9786  \\
    PEM & \textbf{0.8844 } & 0.8399  & \underline{0.8671}  & 0.8110  \\
    SAD   & 0.9932  & \underline{0.9945}  & 0.9914  & \textbf{0.9977 } \\
    SRS1  & \textbf{0.9249 } & \underline{0.8737}  & \underline{0.8737}  & 0.8430  \\
    SRS2  & \textbf{0.5778 } & \textbf{0.5578}  & 0.5056  & \underline{0.5389}  \\
    \midrule
    Average & \textbf{0.8764}  & \underline{0.8576}  & 0.8538  & 0.8421  \\
    \bottomrule
    \end{tabular}%
    }
  \label{tab:alignment}%
\end{table}%

\subsubsection{Effect of Alignment Strategies. } 

To investigate the impact of different alignment strategies, this study examines the classification accuracy of four variants across multiple datasets. As shown in Table \ref{tab:alignment}, the experimental results demonstrate that the coarse-grained and fine-grained alignment strategy outperforms the other strategies on most datasets, with an average accuracy significantly higher than the others, while the Random alignment strategy achieves the lowest average accuracy. This indicates that dynamic time series features are effectively integrated and utilized under this strategy, thereby providing a consistently substantial improvement in the model’s classification performance across a variety of scenarios and usage contexts. Notably, on the SAD dataset, the fine-grained alignment strategy surpasses the coarse-grained and fine-grained alignment, suggesting that the feature structure of this dataset is relatively simple and stable. In this case, fine-grained alignment is more effective in capturing key features to improve classification accuracy, whereas coarse-grained alignment does not provide any additional benefit. For the other datasets, the combined coarse-grained and fine-grained alignment strategy demonstrates stronger overall performance, generally outperforming any single alignment method and delivering more reliable behavior across diverse conditions. In the future, the alignment strategy can be further optimized to better adapt to the specific characteristics of different datasets, potentially enhancing the model’s overall performance and ensuring even greater flexibility and stability across tasks.

\begin{figure}[h]
    \centering
    \includegraphics[width=0.65\linewidth]{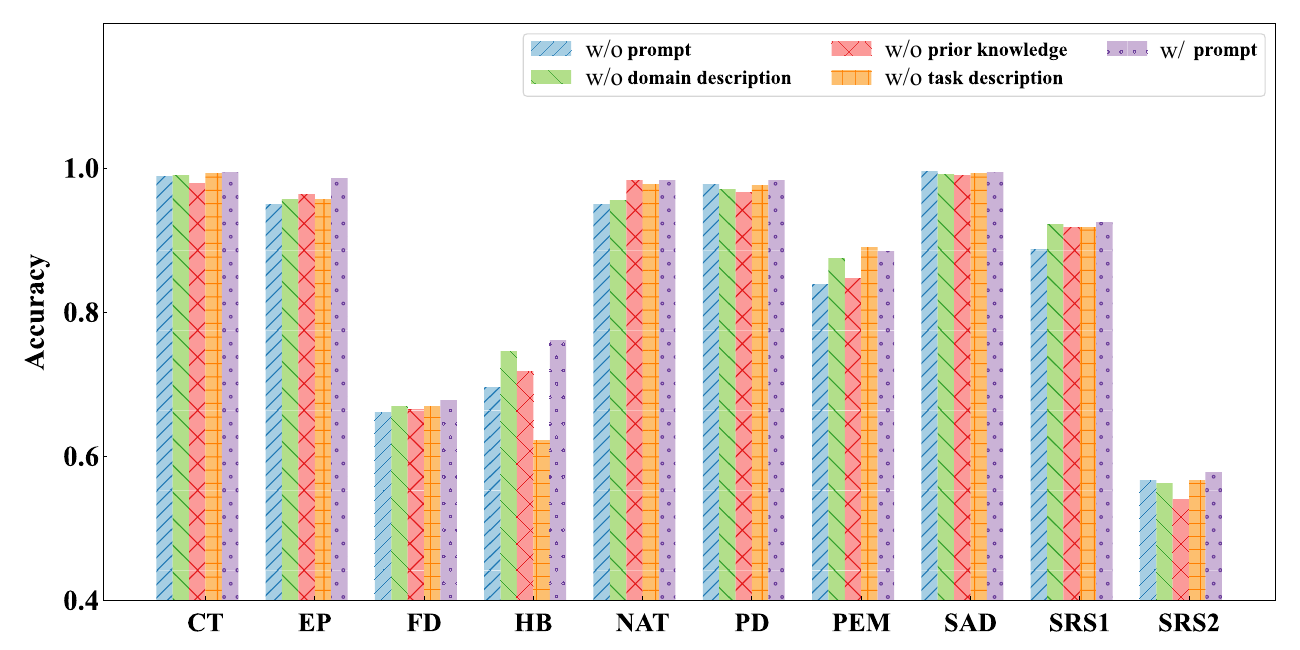}
    \caption{Comparison of the effects of different prompt components on model accuracy across 10 datasets.}
    \label{fig:prompt}
\end{figure}

\subsubsection{Impact of Prompt Design Strategy. }

To analyze the influence of different cue components on model performance, this study systematically evaluates how the removal of prior knowledge, task description, and domain description affects classification accuracy across multiple datasets. As shown in Figure \ref{fig:prompt}, prior knowledge and task description exert a notably more significant impact on overall performance. Compared to the complete prompt configuration, the model’s accuracy drops substantially when these two components are excluded, particularly on the HB dataset, indicating their critical role in enhancing model effectiveness.
In contrast, the removal of domain description results in a comparatively smaller decline in accuracy, suggesting that its influence on overall model performance is relatively weaker. Therefore, prior knowledge and task description emerge as the most influential components, and their absence leads to a pronounced reduction in accuracy. Under the w/o prompt condition, the model’s accuracy is consistently lower, especially on the NAT and SRS1 datasets, further reinforcing the importance of structured prompt design.
These findings emphasize that a well-designed prompt is essential for optimizing model performance. Overall, the prompt design strategy significantly enhances classification accuracy, with the model performing considerably better when the complete prompt configuration is utilized, compared to cases where only individual components are incorporated.

\begin{figure}[h]
    \centering
    \includegraphics[width=0.65\linewidth]{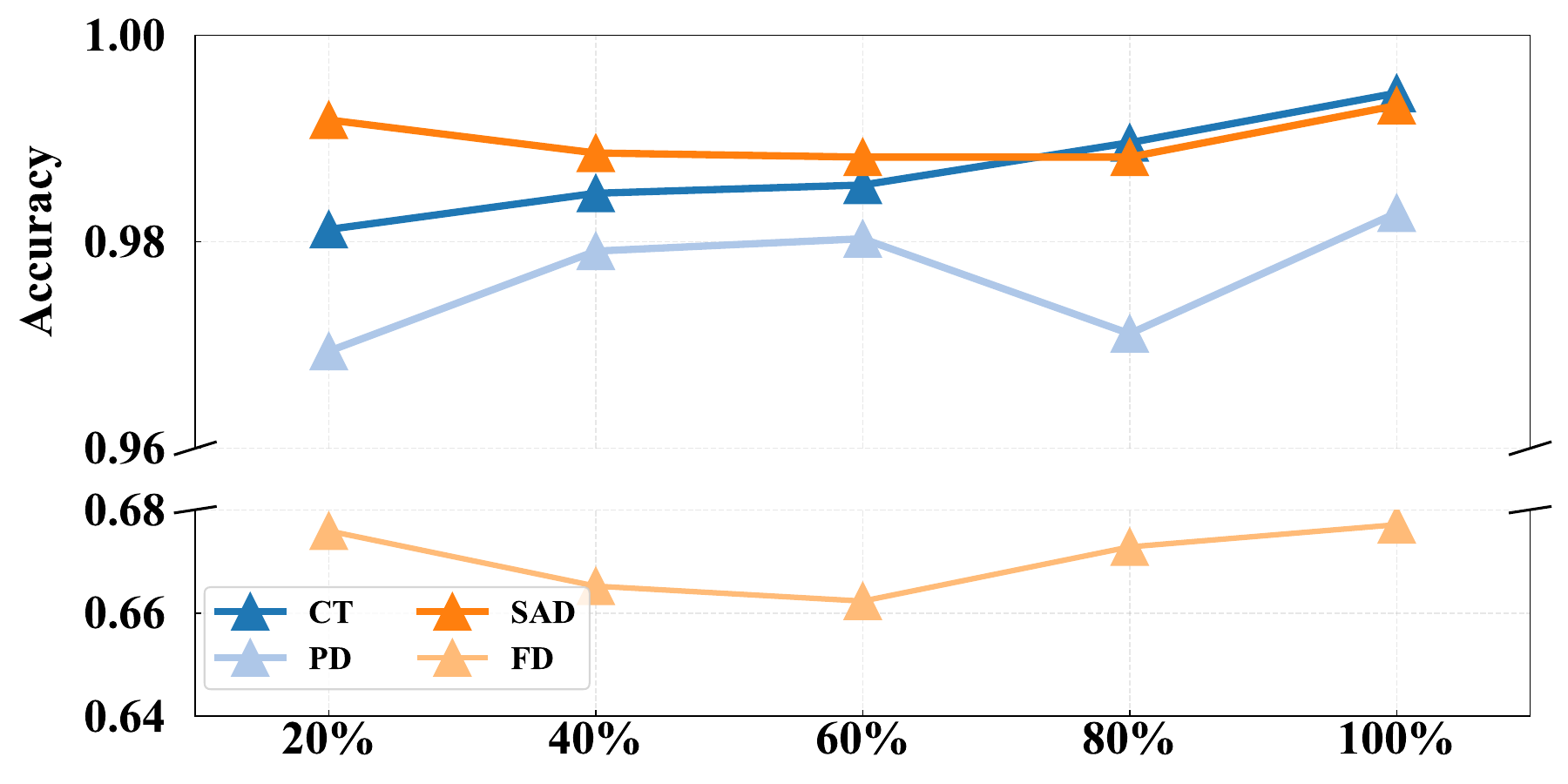}
    \caption{Few-shot performance evaluation of HiTime under varying training data ratios (20\%–100\%) across multiple time series classification benchmarks. }
    \label{fig:few_shot}
\end{figure}

\subsection{In-depth Analysis}
\subsubsection{Few-shot Evaluation}
In this section, we examine HiTime’s performance under few-shot scenarios with varying amounts of training data across multiple datasets. As shown in Figure~\ref{fig:few_shot}, HiTime maintains relatively stable performance across datasets without a substantial accuracy drop, demonstrating its stability under data-scarce conditions. On the SAD dataset, HiTime sustains high accuracy across all sample sizes, suggesting that its hierarchical feature extractor effectively captures dynamic and well-structured temporal patterns even with limited supervision. In contrast, the CT dataset exhibits a noticeable performance decline as the number of training samples decreases, reflecting HiTime’s sensitivity to data scale and its reliance on sufficient examples to achieve stable alignment between semantic and temporal features. On the PD dataset, HiTime performs the weakest, particularly in few-shot settings, where the high intra-class variability and irregular temporal transitions hinder consistent feature generalization. Nevertheless, the overall stability across other datasets indicates that HiTime possesses a strong capacity to adapt to low-data regimes while preserving temporal–semantic coherence.

\subsubsection{Inference Efficiency Analysis}
Table \ref{tab:infer_speed} compares the inference efficiency of different models on a GPU with a batch size of 32. ConvTimeNet achieves the highest throughput and iteration speed due to its lightweight convolutional architecture, which, however, comes at the cost of limited representational capacity. Among LLM-based models, HiTime demonstrates superior overall efficiency, achieving a throughput of 210 samples/s while maintaining lower memory consumption compared with Time-LLM and Time-FFM. This indicates that HiTime provides a more balanced trade-off between computational efficiency and model complexity. The improvement can be attributed to its hierarchical feature extractor and alignment mechanism, which enhance information utilization without introducing substantial inference overhead. Although GPT4TS achieves slightly higher throughput than HiTime, the gap is marginal, and HiTime offers better memory efficiency and stronger classification performance.

\begin{table}[t]
\centering
\small
\caption{Comparative inference efficiency on GPU (batch size = 32).}
\begin{tabular}{lccc}
\toprule
Models & Throughput (samples/s) $\uparrow$ & Iter/s (batch=32) $\uparrow$ & Memory (GB) $\downarrow$ \\
\midrule
ConvTimeNet & 1,280 & 40.0 & 1.4 \\
GPT4TS & 230 & 7.2 & 14.8 \\
Time-LLM & 190 & 6.0 & 16.5 \\
Time-FFM & 175 & 5.6 & 15.9 \\
HiTime (Ours) & 210 & 6.8 & 15.2 \\
\bottomrule
\end{tabular}
\label{tab:infer_speed}
\end{table}

\begin{figure}[h]
    \centering
    \includegraphics[width=0.8\linewidth]{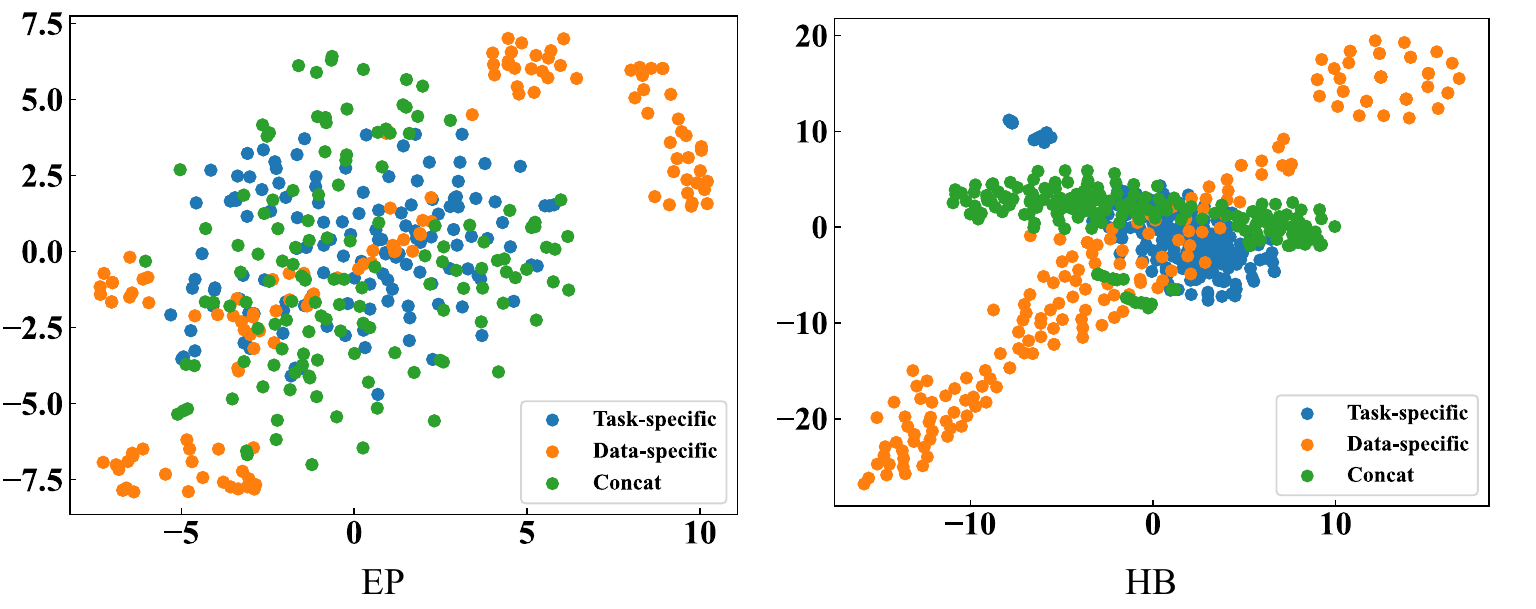}
    \caption{t-SNE visualization of the representations extracted by different encoders on two datasets. The x- and y-axes correspond to the two principal dimensions of the t-SNE embedding space, where samples with similar underlying characteristics are projected closer together.}
    \label{fig:t-sne}
\end{figure}

\subsection{Case Studies}
\subsubsection{Visualization of Feature Extraction.}
To visualize the impact of different encoding strategies on feature representation, we map the features from various datasets and compare those extracted using the three encoding strategies, as shown in Figure \ref{fig:t-sne}. Data-specific features capture intrinsic dataset-level patterns, while task-specific features focus on task-relevant discriminative cues, resulting in a more compact and separable distribution. The concatenated representation integrates both perspectives and reflects a richer feature space.
In the EP dataset, the concatenated features closely align with task-specific features, indicating that the model emphasizes task-relevant information. This alignment leads to performance comparable to the task-specific encoder while surpassing the data-specific encoder. In the HB dataset, the concatenated features incorporate complementary information from both feature types, enabling a more comprehensive representation and outperforming single-encoder strategies under this more heterogeneous setting.

\subsubsection{Effect of Text and Time Series Placement.}
\begin{figure*}[t]
  \centering
    \subfloat[Prompt before time series aligned input]{
    \label{fig:prompt_before}
    \includegraphics[width=.49\linewidth]{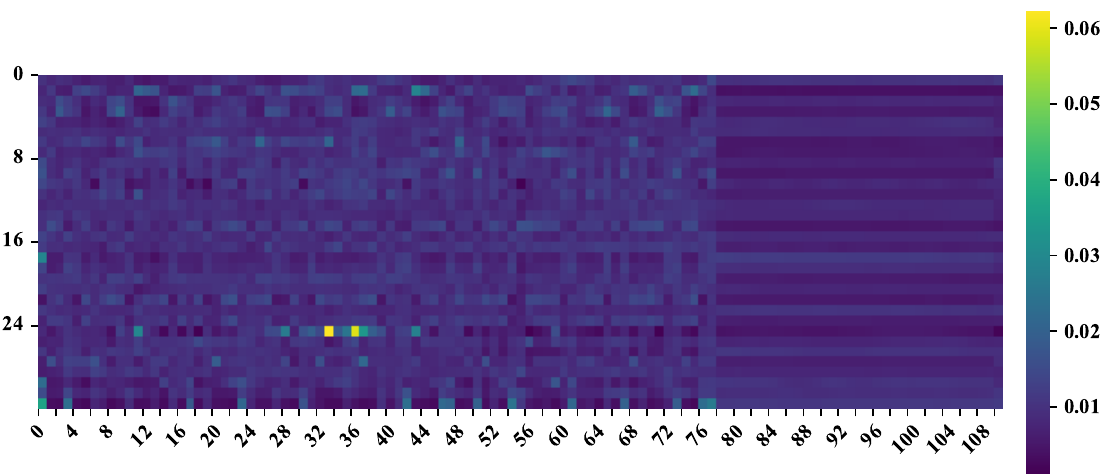}
	}
   \subfloat[Prompt after time series aligned input]{
    \label{fig:prompt_after}
    \includegraphics[width=.49\linewidth]{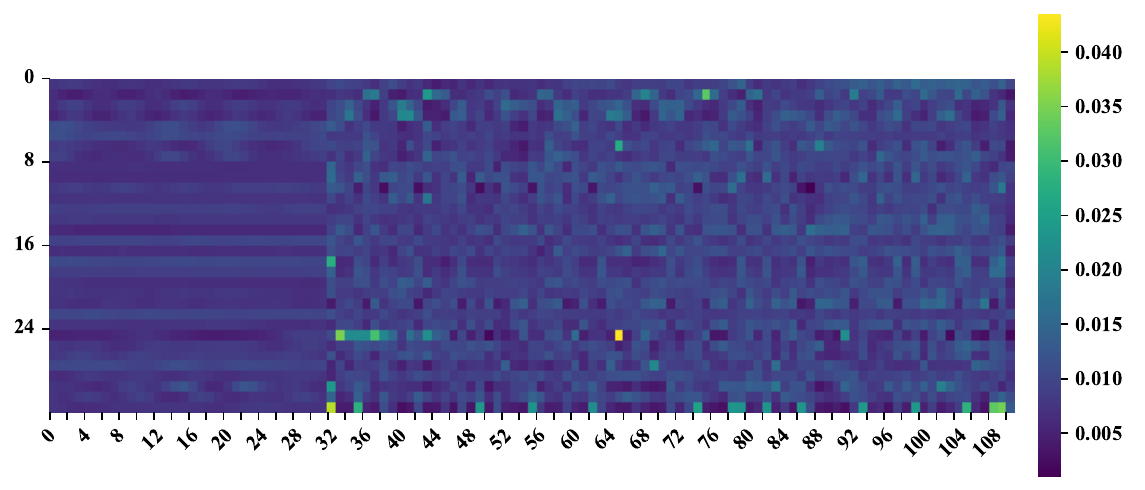}
	}
  \caption{Analysis of the impact of time series and text alignment on LLMs attention distribution. The x-axis represents the embeddings of both text and time series data, while the y-axis shows the corresponding attention distribution across different attention heads. }
    \label{fig:prompt_pos}
\end{figure*}

In this section, we visualize the attention distribution of the LLM when the textual prompt is placed before and after the aligned time series input. As shown in Figure \ref{fig:prompt_pos}, the model tends to allocate relatively higher attention to textual tokens, reflecting its strong linguistic prior learned during pre-training and its inherent preference for processing semantic information first. This observation does not indicate a failure of the proposed framework but rather highlights an intrinsic bias of LLMs toward text-dominant representations. Such behavior further underscores the importance of our alignment design, which aims to guide the model to better integrate temporal cues with semantic context instead of relying solely on linguistic features.
By placing the prompt after the time series, the model first encodes the multimodal patterns contained in the aligned temporal input and then interprets the textual instructions within that context. This ordering encourages the LLM to attend to temporal structures more faithfully, leading to improved contextual alignment and more accurate downstream predictions. The resulting attention distribution demonstrates that prompt positioning plays a meaningful role in shaping how the model balances temporal and semantic information during inference.

\section{Conclusion and Limitations}
In this work, we presented HiTime, a hierarchical LLM-based framework that reformulates time series classification as a generative inference problem by bridging structured temporal representations and language-based semantic reasoning. By integrating a hierarchical sequence feature encoding module, a semantic space alignment mechanism, and a parameter-efficient supervised fine-tuning strategy, HiTime enables large language models to effectively capture temporal dynamics while maintaining semantic consistency across modalities. The proposed framework provides an effective solution for multimodal time series classification and demonstrates strong potential in applications that require both accurate temporal modeling and high-level semantic interpretation. Extensive experimental results on multiple benchmarks validate the effectiveness and generality of the proposed approach.

Despite its encouraging performance, this study also has several limitations. HiTime relies on LLMs as the semantic backbone, which may introduce computational overhead in resource-constrained settings. Moreover, HiTime is not a foundation model, which limits its generality and scalability. It also lacks long chain-of-thought reasoning capability, restricting its ability to perform deeper temporal reasoning on complex sequences. In addition, its effectiveness may depend on prompt quality and the availability of paired multimodal data. Future work will explore more efficient architectures, improved prompt optimization, and extensions to broader time series tasks.
\section*{Acknowledgments}

This research was supported by grants from the National Natural
Science Foundation of China (No. 62502486, 62337001), the grants of the Provincial
Natural Science Foundation of Anhui Province (No. 2408085QF193), USTC Research Funds of the DoubleFirst-Class Initiative (No. YD2150002501),
the Fundamental Research Funds for the Central Universities of
China (No. WK2150110032).
\bibliographystyle{ACM-Reference-Format}

\bibliography{main}

\end{document}